# HoloPOCUS: Portable Mixed-Reality 3D Ultrasound Tracking, Reconstruction and Overlay

Kian Wei Ng[1,2][0000-0002-3485-1191], Yujia Gao[1], Shaheryar Mohammed Furqan[1], Zachery Yeo[1], Joel Lau[1], Kee Yuan Ngiam[1] and Eng Tat Khoo*[2][0000-0003-1295-3506]

[1] National University Health System, S119224, Singapore
[2] College of Design and Engineering, National University of Singapore, S117575, Singapore
kianwei@u.nus.edu
*etkhoo@nus.edu.sg

**Abstract.** Ultrasound (US) imaging provides a safe and accessible solution to procedural guidance and diagnostic imaging. The effective usage of conventional 2D US for interventional guidance requires extensive experience to project the image plane onto the patient, and the interpretation of images in diagnostics suffers from high intra- and inter-user variability. 3D US reconstruction allows for more consistent diagnosis and interpretation, but existing solutions are limited in terms of equipment and applicability in real-time navigation. To address these issues, we propose HoloPOCUS — a mixed reality US system (MR-US) that overlays rich US information onto the user's vision in a point-of-care setting. HoloPOCUS extends existing MR-US methods beyond placing a US plane in the user's vision to include a 3D reconstruction and projection that can aid in procedural guidance using conventional probes. We validated a tracking pipeline that demonstrates higher accuracy compared to existing MR-US works. Furthermore, user studies conducted via a phantom task showed significant improvements in navigation duration when using our proposed methods.

## 1 Introduction

Modern medical imaging provides essential information for diagnostics and intervention. CT and MRI provide 3D anatomical information but exposes users to ionizing radiation and are not suitable for patients with ferrous implants respectively [1]. Ultrasound (US) imaging provides a relatively low-cost, mobile, and safe alternative [2], but in the conventional 2D form the results require more experience to interpret. This impacts diagnostic power as well as intervention efficacy. Studies have shown that using 2D US for diagnosis suffers from high inter-user variability [3], and effective intervention using 2D US is correlated with clinical experience and training [4].

Existing works have been proposed to address some of the limitations of conventional 2D US. 3D US volumes can be either captured directly via specialized probes [5], or reconstructed by stitching individual frames into a volume. Volumetric 3D US reconstruction requires an estimation of the relative pose between frames, with ap-proaches involving electro-magnetic [6], IMU [7], and sensor-less deep learning being proposed [8]. Inter-user diagnostic variability has been shown to improve with the usage of 3D US volumes and the associated features [9].

Using medical imaging for intra-operative or interventional procedure guidance allows clinicians to navigate to or around anatomy not visible to the naked eye due to tissue occlusion [10]. The direct fusion of imaging data onto the user's vision, superimposed onto the actual anatomy, can provide a more intuitive and usable system that could improve the accuracy and speed of procedures [11]. Several works have leveraged on mixed-reality (MR) hardware, proposing to superimpose point-of-care US slices onto the user's vision, reducing the cognitive load required for clinicians to register and reproject the images onto the body [12–16].

**Table 1.** Summary of related works in MR-US.

| | Tracking Method | Hardware | Projection/Overlay |
|---|---|---|---|
| [13] | Opto-electronic | 2D Probe; HMD; tracking equipment | Image |
| [12] | Electro-magnetic | 2D Probe; Monitor; tracking equipment | Image |
| [14] | Monocular + ArUco | 2D Probe; HMD | Image |
| [15] | Depth/IR + Spheres | 2D Probe; HMD | Image |
| [16] | Stereo + Spheres | 3D Probe; HMD | Volume |
| Ours | Stereo + ArUco | 2D Probe; HMD | Image / Volume |

For the US slices to be registered and overlaid onto the body, the US probe needs to be tracked; MR-US solutions such as [13] and [12] utilized specialized tracking equipment such as opto-electronic or electromagnetic systems. While benefiting from high accuracy, the additional hardware adds to cost and reduces portability. [14] and [15] instead used cameras on head-mounted devices to directly track the probe, using fiducial markers with monocular and Infrared (IR)/Depth feeds respectively. With the acquisition, tracking and projection system integrated into one device, the need for additional equipment is removed. While portability is improved, neither have validated tracking results that are close to clinically requirements [10]. While most prior works focus on the visual overlay of 2D slices, [16] utilizes specialized probes that directly acquire and project 3D volumetric data in contrast to conventional 2D probes. We advance the domain and application of MR-US with the following contributions:

- We developed a stereo-tracking pipeline that extracts richer fiducial keypoints, which can be filtered and processed to provide higher accuracy tracking and MR-US 2D overlay compared with existing works.
- Existing solutions that utilize conventional linear probes enable the visual overlay of 2D US slices. Our proposed system allows for users to reconstruct and project 3D MR-US data, to be used in direct intervention or downstream diagnostic tasks.
- We conducted a user study to test the effectiveness of both 2D and 3D MR-US solutions against conventional US operation for a simulated biopsy task, providing insights into the potential benefits and drawbacks of implementing such systems in different (e.g. diagnostic/interventional) clinical settings.

# 2 Methods

## 2.1 System Architecture

HoloPOCUS utilizes Microsoft's HoloLens 2 for sensing and visualization [17]. HoloLens 2 provides multiple cameras – one high definition RGB, four grayscale with overlapping field-of-view (FOV), and an IR/Depth. From **Table 1**, [14] and [15] that use the HoloLens line of device utilized RGB and IR/Depth respectively. While the RGB feed provides high resolution images, the FOV does not cover the typical region used for tracking hand-held objects (Supp. Fig. 1) [18]. Conversely, the IR/Depth feed has a wide FOV but suffers from accuracy issues related to both random and warm-up variability [19, 20]. Given the above hardware limitations, we opted to use the stereo streams (Supp. Fig. 1) [18], with the benefit of a FOV that includes hand-object interactions, and high accuracy and reliability stemming from stereo triangulation.

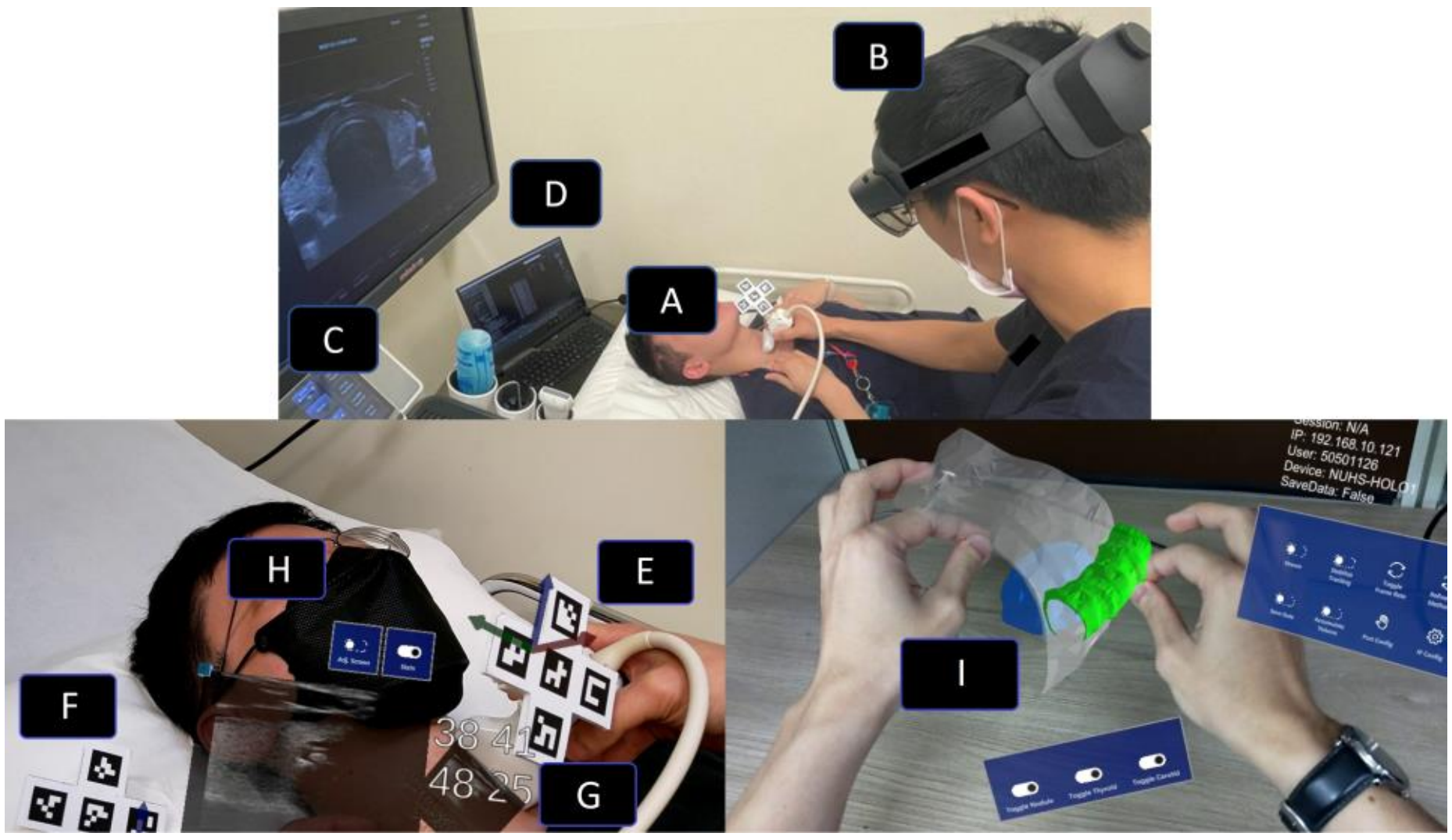

**Fig. 1. (Top)** Clinician using HoloPOCUS; **(Bottom Left)** First person view with 2D overlay (G), large virtual screen for viewing fine detail (H), operating distance/angle as user feedback; **(Bottom Right)** Tracking of US slices over time allows for 3D reconstruction of nodule and surrounding structures e.g. carotid/thyroid (I), which can be projected directly back on the acquisition location or inspected post-hoc (as shown). 150 slices were used for reconstruction.

A custom 3D-printed attachment was made to secure ArUco markers to the probe for tracking. The attachment was designed with two joints that can be rotated at 45° intervals, providing greater flexibility compared to [14] for probe positioning and orientation that can differ significantly based on anatomy and procedures.

To track and project US data onto the user's vision, US images are streamed from the US machine **(Fig. 1C)** to a laptop for processing **(Fig. 1D)**. Simultaneously, the stereo feed from HoloLens **(Fig. 1B)** is streamed to the laptop to compute the fiducial markers' pose. Since the markers **(Fig. 1A)** are placed at a known offset from the probe tip, an offset transformation is applied to compute the pose of the probe tip. The pose is then paired with the US data for rendering in the user's vision. By tracking the slices across space-time **(Fig. 1, Bottom Left)**, we demonstrate the ability to reconstruct the 3D anatomy for richer visualization/guidance **(Fig. 1, Bottom Right)**.

### 2.2 Dense Fiducial Keypoints Extraction for Stereo Pose Estimation

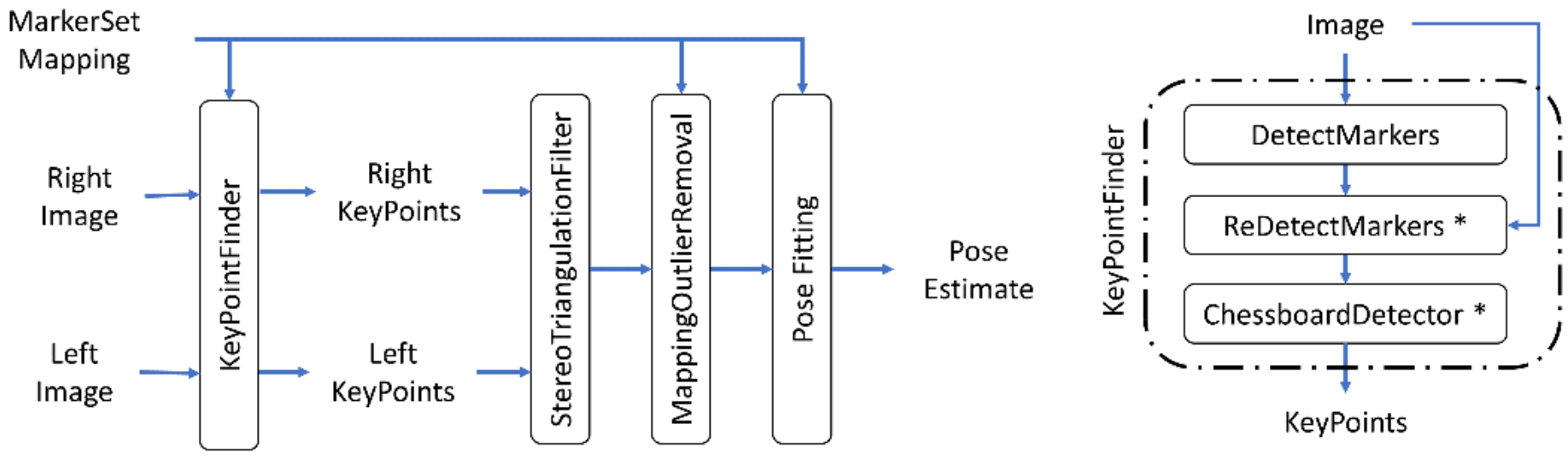

**Fig. 2.** Stereo pairs are processed independently with the KeyPointFinder sub-module (* denotes steps requiring MarkerSet Mapping), followed by triangulation, filtering, and pose fitting.

Two stages are applied to retrieve the marker pose from a stereo pair. ArUco markers are identified in each image (*DetectMarkers*) [21]. A secondary detection pass is done on the image (*ReDetectMarkers*) to detect any previously missed markers, using the known MarkerSet mappings as reference. Given an $n$ marker configuration, only up to $4n$ corners can be extracted. Prior works augmented ArUco for more keypoints either by adding features [22–24] or densely predicting keypoints in the binary pattern via a GPU-based deep learning approach [25]. In this stage (*ChessboardDetector*), we exploit natural chessboard corners found in ArUco patterns. This targeted approach reduces computation and gives us extra high-quality keypoints due to the well-defined intersections provided by chessboard corners [26], with the same spatial footprint.

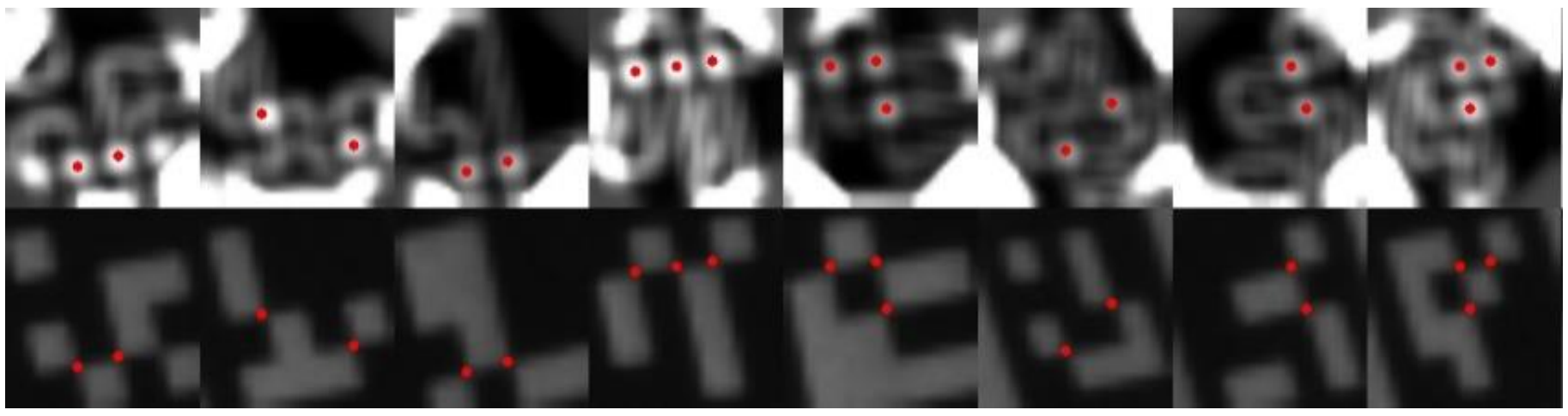

**Fig. 3.** Filter response for cropped ArUco patches, with local maxima filtered for points that are close to guesses interpolated from the original 4 corners.

For each ArUco marker detected, we crop and upsample the patch to a constant size. A radon-based transform was used to extract the response map (**Fig. 3, top row**) [26]. Given the original 4 corners for each ArUco, we interpolated to extract candidate guesses for where chessboard corners could be located. Local maxima from the response map are matched to these guesses (**Fig. 3, red points**) and refined via a weighted average of the local response.

$$s_i = \frac{1}{c_1 - 1} \sum_{j \neq i} |d(i,j) - r(i,j)| \quad (1)$$

With the enlarged keypoint sets found for both left and right images, stereo matching is done with the camera intrinsic, with stereo rays that do not intersect within a fixed tolerance (1mm) being discarded. We perform an outlier removal step on the remaining $c_1$ 3D points. Pairwise distances $d(i,j)$ are computed exhaustively and compared against the ground truth reference $r(i,j)$ **(Eq. 1)**. The score $s_i$ is computed for each point, with those above a fixed threshold (0.75mm) being discarded.

The resulting $c_2$ points each have a confidence value assigned from stereo intersection. We compute $m$ candidate poses using the top $\{c_2 - 1, c_2 - 2, \ldots, c_2 - m\}$ confidence points, retaining the pose with the lowest fiducial registration error (FRE).

### 2.3 Projection Computation

This section describes the integration of stereo tracking output with coordinate systems across devices, for 2D slice projection or 3D reconstruction-projection tasks.

**Real-time 2D slice reprojection.** The tracking module returns the computed 4x4 transformation matrix ${}^{P}_{C}P$ that provides the pose of the probe's ArUco marker set in relation to the cameras. We retrieve the computed pose of the camera relative to a static world coordinate system ${}^{C}_{W}P$ using the ResearchMode API [17]. For the US slices to be projected at the correct location within the patient's body, we precompute another transformation ${}^{T}_{P}P$ that describes the marker's relation to the probe tip.

$${}^{T}_{W}P = {}^{C}_{W}P * {}^{P}_{C}P * {}^{T}_{P}P \quad (2)$$

Chaining these transformations **(Eq. 2)** allows us to retrieve ${}^{T}_{W}P$, the final transformation relating the position of the US slice at the probe tip relative to the application's world coordinate system. This pose is computed on a per-frame basis and sent to the headset for real-time projection and rendering of the 2D slices.

**3D reconstruction-reprojection.** US frames that are tracked in a consistent coordinate system over time can be accumulated into a 3D volume. Given the pixel-spacing ($p_w$, $p_h$) mm for the US image, a 4x4 matrix ${}^{I}_{T}P$ is pre-computed via CAD software to transform each pixel's coordinate (x,y,0) to be expressed relative to the tip in 3D.

To improve the accuracy of the 3D reconstruction process, we included an optional "anchor" marker (AM) set **(Fig. 1F)**, identical in design to the probe tracking set **(Fig. 1E)**. Previous works have shown that HoloLens' self-localization via its internal algorithm had an average error of 1-3cm in an indoor-mapping task [20]. While this value fluctuates depending on the environment, we mitigate this source of error by introducing the AM set. By running the tracking module in parallel to track the AM, the transformation matrix ${}^{A}_{C}P$ containing the pose of AM relative to the camera.

$$ {}^{I}_{A}P = {}^{A}_{C}P^{-1} * {}^{P}_{C}P * {}^{T}_{P}P * {}^{I}_{T}P \quad (3) $$

$$ {}^{I}_{W}P = {}^{C}_{W}P * {}^{P}_{C}P * {}^{T}_{P}P * {}^{I}_{T}P \quad (4) $$

Combining the terms through **Eq. 3** gives us ${}^{I}_{A}P$, a new way of expressing the pixel data with respect to AM's coordinate system.

Let $E_{stereo}$ and $E_{local}$ represent the errors present in pose computation via the stereo tracking module and HoloLens' self-localization respectively. In terms of error contribution, accumulating the data relative to the world (**Eq. 4**, ${}^{I}_{W}P$) would result in $E_{recon|W} = E_{stereo} + E_{local}$, stemming from ${}^{P}_{C}P$ and ${}^{C}_{W}P$ respectively. On the other hand, accumulating the data relative to AM (**Eq. 3**, ${}^{I}_{A}P$) would result in $E_{recon|A} = E_{stereo} + E_{stereo}$, stemming from ${}^{P}_{C}P$ and ${}^{A}_{C}P$ being tracked independently.

Given the above, AM should be used for reconstruction if $E_{stereo} \ll E_{local}$. With an evaluation of how usage parameters affect $E_{stereo}$, an upper-bound for $E_{recon|A}$ can be estimated, which would not be possible in the case of $E_{recon|W}$ due to the unpredictable nature of the $E_{local}$ component.

### 2.4 Implementation Details

We utilized a laptop (i9-12900H CPU) for computation together with a linear probe (Mindray DC-80A, L14-5WE) for expert-user testing and feedback. The HoloLens application was developed with Unity, gRPC and MRTK. We used the system with a wireless probe (SonoStarMed, 128E) that streamed data to an iPhone 12 mini for the user study. All processing ran in real-time for a 30 Hz HoloLens stereo feed, with full keypoint extraction and pose estimation averaging **25.6** and **5.2 ms** respectively.

## 3 Results

### 3.1 Tracking Accuracy

Of the related MR-US works described **(Table 1)**, we excluded [16] from the tracking comparison as they utilized a unique setup (1920x1080 high definition RGB stereo) that is not available to HoloLens 2 and had reported FRE metrics. For MR-guided navigation, FRE has been shown to be uncorrelated with overlay accuracy [27, 28].

**Table 2.** Tracking results comparison. * Indicates results reproduced with experimental setup changes as described. Results with inclusion of chessboard keypoints indicated in brackets.

| | Tracking Method | Translation RMS | Rotation RMS |
|---|---|---|---|
| [14] * | Mono, 2x ArUco | 13.8 mm | 10.9 ° |
| [14] * | Mono, 5x ArUco | 10.9 mm | 7.91 ° |
| [15] | Depth/IR, Spheres | 2.81 mm | 1.70 ° |
| [29] | Stereo, Spheres | 1.90 mm | 1.18 ° |
| [29] | Mono, 1x ArUco | 6.09 mm | 6.73 ° |
| Ours | Stereo, 2x ArUco (+C) | 1.08 mm (0.91 mm) | 1.31 ° (1.24 °) |
| Ours | Stereo, 5x ArUco (+C) | 0.49 mm (0.45 mm) | 0.60 ° (0.60 °) |

Instead, we focus on works utilizing HoloLens 2 for fiducial tracking [15, 29], with sensors that had a suitable FOV. For [14], we utilized the original monocular PnP estimation on ArUco markers setup but opted to use the wide FOV grayscale 640x480 instead of the original low FOV RGB 1920x1080 feed for a fair FOV comparison.

For evaluation, past works moved markers along a known trajectory, with frame-to-frame poses compared against a gold standard. We simulated this by placing two sets of markers at a known offset. The viewing distance and angles were varied, with relative poses for each set computed per frame and compared against this offset.

We show that even with a 2 ArUco marker configuration (minimum of 2 markers needed for *ReDetectMarkers* module), our pose translation and rotation errors outperform existing solutions. The inclusion of chessboard corners had a stronger effect on low marker setups, with negligible improvements when 5 markers are used.

**Table 3.** Effect of usage distance and angle to marker on translation and rotational RMS.

| | | Usage Angle. (deg) | | |
|---|---|---|---|---|
| | | 0-15 | 15-30 | 30-45 |
| Usage Dist. (cm) | 15-25 | 0.222 mm / 0.417 ° | 0.228 mm / 0.407 ° | 0.187 mm / 0.459 ° |
| | 25-35 | 0.284 mm / 0.288 ° | 0.282 mm / 0.554 ° | 0.288 mm / 0.441 ° |
| | 35-45 | 0.561 mm / 0.737 ° | 0.621 mm / 0.756 ° | 0.631 mm / 0.675 ° |
| | 45-55 | 0.859 mm / 0.912 ° | 0.854 mm / 0.950 ° | 1.015 mm / 0.961 ° |

For effective and reliable usage of HoloPOCUS, we investigated the effect of the cameras' distance and angle relative to the markers on accuracy **(Table 3)**. Within the defined operational limits for the 5-marker configuration, translation and rotation RMS ranged from **0.19-1.02 mm** and **0.41-0.96°** respectively.

Lastly, the experimental setup allowed us to track and compute ${}^{P}_{W}P$ and ${}^{P}_{A}P$ simultaneously. We estimated $E_{local}$, the variation in probe tracking due to self-localization uncertainty to be around **1-2 mm**, in line with past experimental results [20].

### 3.2 User Study

To evaluate HoloPOCUS' effectiveness, we compared it against conventional US for a phantom biopsy task, using the time taken as a quantitative metric. Following [30], sets of three targets were submerged in agar at 10-20 mm depths (Supp. Fig. 2). Each set was contained in a 15x5 cm block, consisting of two small and one large target, with 7.5 and 15 mm diameters respectively. This design followed ATA guidelines for thyroid nodule biopsy [31]. For each trial, users were tasked to use the selected US method to locate and hit the three targets in succession with a needle. The order of methods was randomized to account for task familiarity bias.

We recruited an equal number of novices and experts, with novices defined as individuals with no medical training, and experts as specialists (from specialties that routinely use US as navigational guidance) who have had at least 5 years of post-graduate experience/training. None of the participants had a substantial background in mixed-reality usage. Novices were instructed on the principles of US operation prior to the timed task. The 3D method timing included a reconstruction sweep, which took 18 seconds on average to cover the 15 cm length.

**Table 4.** Time taken (mean ± s.d) and statistical test results for phantom biopsy task.

| | Conventional US | | 2D Overlay | | 3D Recon/Overlay | |
|---|---|---|---|---|---|---|
| | time (s) | - | time (s) | p-value | time (s) | p-value |
| Novices (n=12) | 72.2 ± 43.2 | - | 51.6 ± 19.7 | 0.1838 | 37.1 ± 12.5 | **0.0249** |
| Experts (n=12) | 67.9 ± 27.2 | - | 58.1 ± 25.1 | 0.1253 | 35.0 ± 8.8 | **0.0022** |

A paired two-tailed t-test against conventional US showed a significant reduction in timings for the 3D method. The 2D method showed an insignificant reduction in timing, in line with prior results [14]. Experts performed the task faster than novices on average, except for when the 2D method was used. This is also in line with prior results [14], reflecting how the 2D method did not provide substantial improvements in mental reprojection and instead worsened timings due to technology unfamiliarity.

## 4 Discussion

We introduce a novel MR-US solution for 3D reconstruction-overlay of US data. This done by introducing a high accuracy stereo fiducial tracking pipeline that allows for the reliable accumulation of 2D slices across time to form a 3D volume [32].

The 3D US volume can be used directly for better interventional guidance, as anatomical structures are better perceived in 3D. A user study showed significant improvement in a simulated biopsy task when using a 3D overlay, even with the sweep duration included. We expect sweep time to be insignificant for complex real-world cases, making the benefits more significant. Apart from navigation, the volumes can be reused for diagnostics (e.g. 3D spatial features, nodule temporal progression) [33].

Future work could include using more complex phantoms to accommodate tasks where multiple structures have to be avoided and targeted. Feedback from users included difficulty in estimating phantom target depths. We hypothesize that this could be addressed with more complex phantoms/reconstructions, where the relative locations of structures, aided by mesh occlusions, could provide better 3D perception.

With a larger sample size, analysis can be done to study the effect of age and specialty on MR-US effectiveness. Finally, similar to other works, our measure of accuracy does not account for inaccuracies from the optical system used for visual overlay. A different task design can potentially shed light on this source of error.

The study was approved by the institutional ethics review board (2021/00464) and received support from the Ministry of Education, Singapore, under the Academic

Research Fund Tier 1 (FY2020), and from the National University Health System (NUHSRO/2021/018/ROS+6/EIM-2nd/03).

## Supplementary Materials

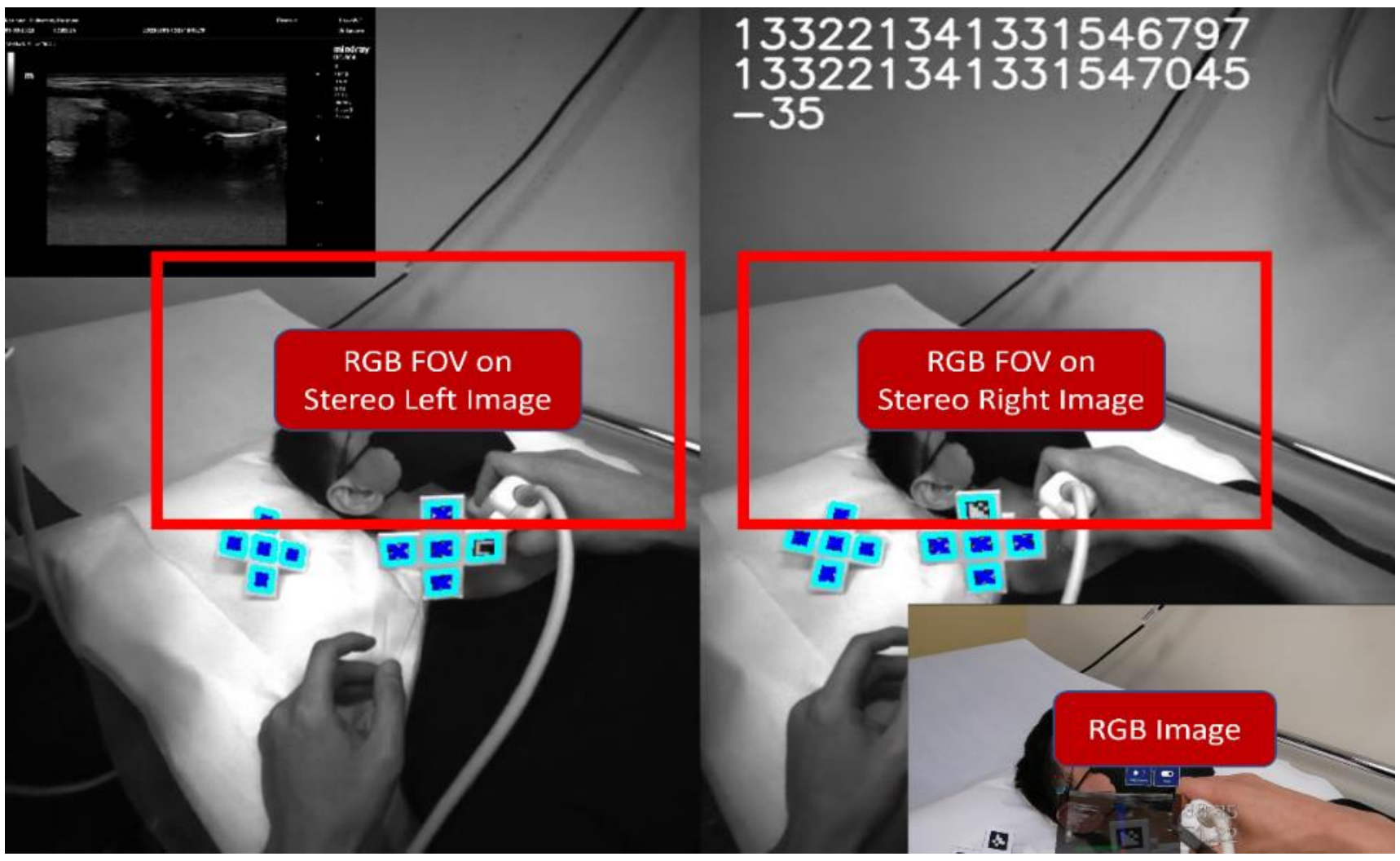

**Supp. Fig. 1.** Stereo and RGB **(bottom right)** feeds for the same scene. Red outlines denote estimated overlaid FOVs, showing the limitation of the RGB feed in capturing hand-object interactions (ArUco markers mostly out of view).

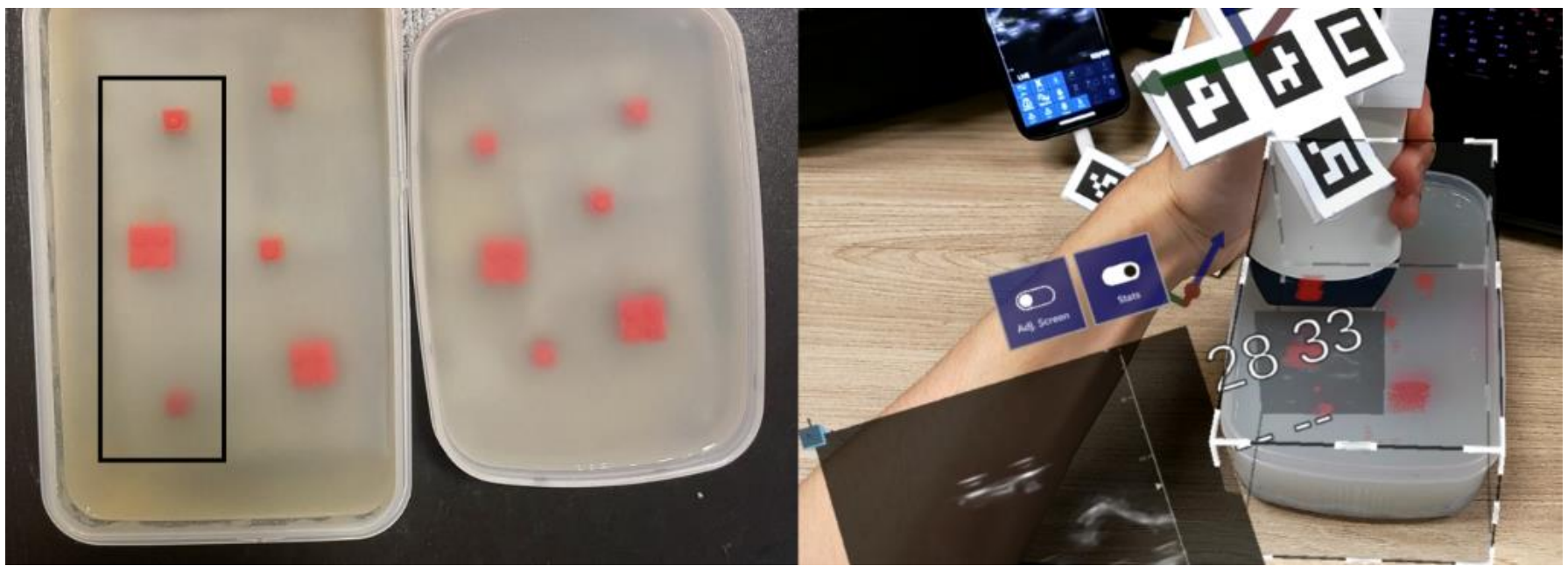

**Supp. Fig. 2. (Left)** Agar phantoms used for user study. Each timed segments require users to locate and hit one vertical column of targets. Targets are submerged at 10-20 mm depths and obscured from the user by a thin opaque sheet that still allows for strong acoustic contact through to the gel. **(Right)** “Conventional” usage entails referencing the iPhone screen, while “2D Overlay” usage users are able to reference the overlaid grayscale slices. The 3D reconstruction sweep creates the red patches, which are then used as reference for the “3D Recon/Overlay” task.